\documentclass[runningheads]{llncs}
\usepackage{graphicx}
\usepackage[T1]{fontenc}
\usepackage[scaled=0.85]{beramono}
\usepackage{listings}
\usepackage{tabularx}
\usepackage[ruled,vlined]{algorithm2e}
\usepackage{float}
 \usepackage{listings, paralist}
 \PassOptionsToPackage{hyphens}{url}
 \usepackage{hyperref}

\usepackage{comment}
\usepackage{todonotes}
\usepackage{wrapfig}
\usepackage[utf8]{inputenc}
\usepackage[T1]{fontenc}

\usepackage{tabularx}

\newcounter{protocol}
\newenvironment{protocol}[1]
  {\par\addvspace{\topsep}
   \noindent
   \tabularx{\linewidth}{@{} X @{}}
    \hline
    \refstepcounter{protocol}\textbf{Protocol \theprotocol} #1 \\
    \hline}
  { \\
    \hline
   \endtabularx
   \par\addvspace{\topsep}}

\definecolor{lightBlue}{RGB}{0, 153, 255}
\definecolor{lightRed}{RGB}{204, 0, 255}

\lstdefinelanguage{Manchester}
{
    sensitive = true,
    keywords = [1]{Class, EquivalentTo, SubClassOf},
    morekeywords = [2]{and, or},
    morekeywords = [3]{some, value},
    keywordstyle=[2]\textbf,
    keywordstyle=[2]\color{lightBlue}\textbf,
    keywordstyle=[3]\color{lightRed}\textbf,
    morestring=[b]''
}

\begin{document}
\title{Explanation Ontology in Action: A Clinical Use-Case
\thanks{Copyright \textcopyright 2020 for this paper by its authors. Use permitted under Creative Commons License Attribution 4.0 International (CC BY 4.0).}
}
%
%
\author{Shruthi Chari\inst{1}\orcidID{0000-0003-2946-7870} \and
Oshani Seneviratne\inst{1}\orcidID{0000-0001-8518-917X} \and
Daniel M. Gruen\inst{1}\orcidID{0000-0003-0155-3777} \and Morgan A. Foreman\inst{2}\orcidID{0000-0002-2739-5853} \and Amar K. Das\inst{2}\orcidID{0000-0003-3556-0844} \and Deborah L. McGuinness\inst{1}\orcidID{0000-0001-7037-4567} }

\authorrunning{S. Chari et al.}

\institute{
Rensselaer Polytechnic Institute, Troy, NY, USA \\
\email{\{charis, senevo, gruend2\}@rpi.edu,dlm@cs.rpi.edu}\\
\and
IBM Research, Cambrdige, MA, USA, 02142 \\
\email{morgan.foreman@ibm.com, amardasmdphd@gmail.com}}
\maketitle              
\begin{abstract}
We addressed the problem of a lack of semantic representation for user-centric explanations and different explanation types in our Explanation Ontology (\url{https://purl.org/heals/eo}). Such a representation is increasingly 
necessary as explainability has become an important problem in Artificial Intelligence with the emergence of complex methods and an uptake in high-precision and user-facing settings. In this submission, we provide step-by-step guidance for system designers to utilize our ontology, introduced in our resource track paper, to plan and model for explanations during the design of their Artificial Intelligence systems. We also provide a detailed example with our utilization of this guidance in a clinical setting.  
\\
\textbf{Resource:} \url{https://tetherless-world.github.io/explanation-ontology}

\end{abstract}
\section{Introduction}
Explainable Artificial Intelligence (AI) has been gaining traction due to increasing adoption of AI techniques in high-precision settings. Consensus is lacking amongst AI developers on the type of explainability approaches and we observe a lack of infrastructure for user-centric explanations. User-centric explanations address a range of users' questions, have different foci, and such variety provides an opportunity for end-users to interact with AI systems beyond just understanding why system decisions were made. In our resource paper \cite{chari2020EO}, we describe an Explanation Ontology (EO), which we believe is a step towards semantic encoding of the components necessary to support user-centric explanations. This companion poster focuses on describing the usage steps (Section \ref{sec:usage}) that would serve as a guide for system developers hoping to use our ontology. We demonstrate the usage of the protocol (Section \ref{sec:clinical}) as a means to support and encode explanations in a guideline-based clinical decision support setting.

\section{Usage Directions for the Explanation Ontology} \label{sec:usage}
\begin{protocol}{Usage of Explanation Ontology at System Design Time}
\textit{Inputs:} A list of user questions, knowledge sources and AI methods \\
\textit{Goal:} Model explanations that need to be supported by a system based on inputs from user studies \\
\textit{The protocol:}
\begin{enumerate}
\item \textbf{Gathering requirements}
\begin{enumerate}
    \item Conduct a user study to gather the user's \textbf{requirements} of the system
    \item Identify and list \textbf{user questions} to be addressed
\end{enumerate}
    \item \textbf{Modeling}
    \begin{enumerate}
       \item Align \textbf{user questions} to explanation types
    \item Finalize \textbf{explanations} to be included in the system
    \item Identify \textbf{components} to be filled in for each explanation type
    \item Plan to populate \textbf{slots for each explanation type} desired
    \item Use the \textbf{structure of sufficiency conditions} to encode the desired set of explanations
    \end{enumerate}
\end{enumerate}
\end{protocol}
System designers can follow the usage directions for EO at design time when planning for the capabilities of an AI system.
The guidance aims to ensure that end-user requirements are translated into user-centric explanations.
This protocol guidance is supported by resources made available on our website. These resources include queries to competency questions to retrieve sample user questions addressed by each explanation type\footnote{\url{https://tetherless-world.github.io/explanation-ontology/competencyquestions/##question2}} and the components to be filled for each explanation type.\footnote{\url{https://tetherless-world.github.io/explanation-ontology/competencyquestions/##question3}} Additionally, the sufficiency conditions that serve as a means for structuring content to fit the desired explanation type can be browsed via our explanation type details page.\footnote{\url{https://tetherless-world.github.io/explanation-ontology/modeling/##modelingexplanations}}
\section{Clinical Use Case} \label{sec:clinical}
We applied the protocol to understand the need for explanations in a guideline-based care setting and identify which explanation types would be most relevant for this clinical use case. \cite{chari2020EO} Further, we utilized the EO to model some of the explanations we pre-populated into the system prototype. Hereafter, we describe our application of the EO usage guidelines and highlight how we used a user study to guide system design. As a part of the user study, we first held an expert panel interview to understand the clinicians' needs when working with guideline-based care. We utilized the expert panel input to design a cognitive walkthrough of a clinical decision support system (CDSS) with some explanations pre-populated at design time to address questions that clinicians would want answered. 

\begin{figure}[hbt!]
\centering
  \includegraphics[width=1.0\linewidth]{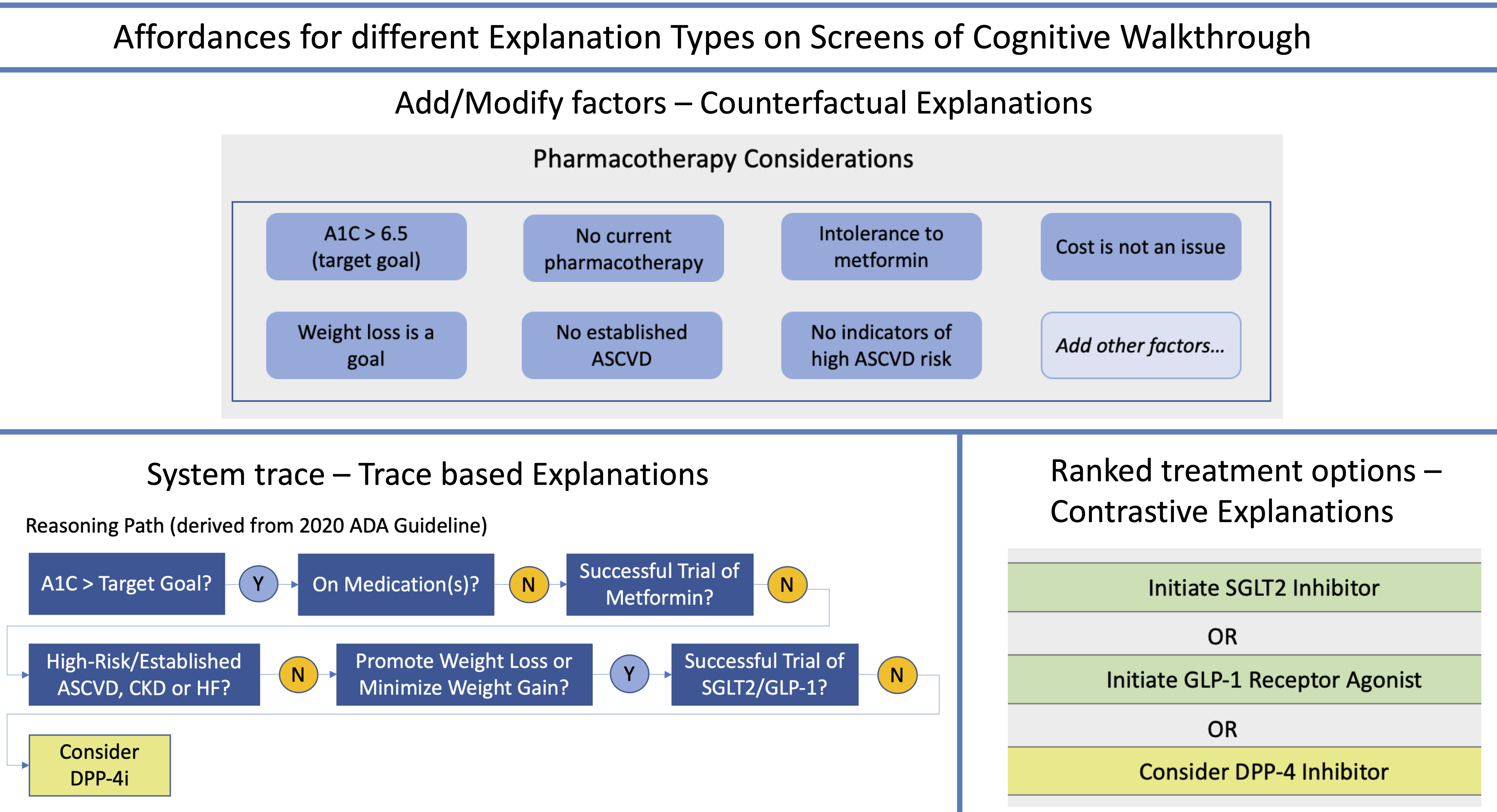}
  \caption{An overview of the user interface affordances we designed into the prototype to accommodate the various explanation types.}
  \label{fig:treatment_planning}
    \vspace{-5mm}
    \end{figure}
    
The prototype of a CDSS we designed for the walkthrough, included allowances for explanations on different screens of the system (Fig. \ref{fig:treatment_planning}), in all, allowing clinicians to inspect a complicated type-2 diabetes patient case. Some examples of the pre-populated explanations are: \textit{contrastive} explanations to help clinicians decide between drugs, \textit{trace based}
explanations to expose the guideline reasoning behind why a drug was suggested, and \textit{counterfactual} explanations that were generated based on a combination of patient factors. Other examples were provided to us by clinicians during the walkthrough. 

\begin{figure}[hbt!]
\centering
  \includegraphics[width=1.0\linewidth]{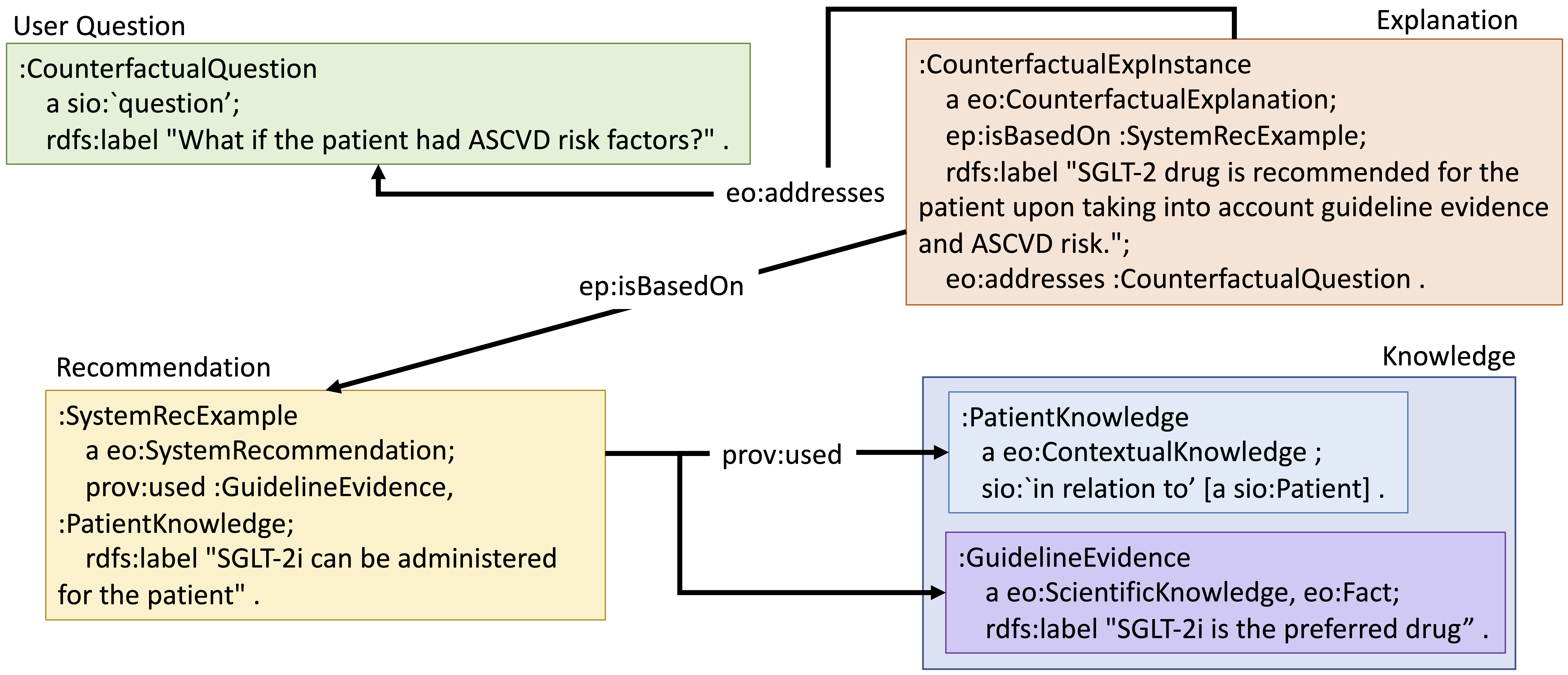}
  \caption{An annotated turtle representation of a counterfactual explanation that the CDSS would output in response to an additional factor the system had not considered.}
  \label{fig:counterfactual}
  \vspace{-5mm}
    \end{figure}

Below, we step-through how the protocol guidance (Section \ref{sec:usage}) can be used to model an example of a counterfactual explanation from this clinical use case. 
A counterfactual explanation can be generated by modifying a factor on the treatment planning screen of the CDSS prototype. 
For illustration's sake, let us suppose that this explanation is needed to address a question, ``What if the patient had an ASCVD risk factor?" where the ``ASCVD risk factor" is an alternate set of inputs the system had not previously considered. 
From our definition of a counterfactual explanation (\url{https://tetherless-world.github.io/explanation-ontology/modeling/#counterfactual}), in response to the modification a system would need to generate a system recommendation based on the consideration of the new input. More specifically, in our use case, a system would need to consider the alternate input of the ASCVD risk factor in conjunction with the patient context and consult evidence from the American Diabetes Association (ADA) guidelines \cite{american20209} to arrive at a new suitable drug recommendation for the patient case. With the alternate set of inputs and the corresponding recommendation in place, the counterfactual explanation components can now be populated as slots based on the sufficiency condition for this explanation type. A Turtle snippet of this counterfactual explanation example can be viewed in Fig. \ref{fig:counterfactual}. Additionally, there have been some promising machine learning (ML) model efforts in the explainability space \cite{arya2019one,ribeiro2016model}  that could be used to generate system recommendations to populate specific explanation types. We are investigating how to integrate some of these AI methods with the semantic encoding of EO.

\section{Related Work}
In the explainable AI space, several research projects \cite{liao2020questioning,wang2019designing,dragoni2020explainable} have begun to explore how to support end-user specific needs such as ours. 
Wang et al. \cite{wang2019designing} present a framework for aligning explainable AI practices with human reasoning approaches, and they test this framework in a co-design exercise using a prototype CDSS. However, their framework isn't in a machine-readable format and hence is difficult to reuse for supporting explanations in a system. The findings from their co-design process that clinicians seek different explanations in different scenarios corroborate with those from our cognitive walkthrough, and EO can help support system designers in this endeavor. Similarly, Dragoni et al. \cite{dragoni2020explainable} propose a rule-based explanation system in the behavior change space capable of providing trace based explanations to end-users to encourage better lifestyle and nutrition habits. While this explanation system was tested with real target users and subject matter experts, it is limited in scope by the types of explanations it provides. Vera et al. \cite{liao2020questioning} have released a question bank of explanation question types that can be used to drive implementations such as ours.

\section{Conclusion}
We have presented guidelines for using our Explanation Ontology in user-facing settings and have demonstrated the utility of this guidance in a clinical setting. This guidance 
describes an end-to-end process to support the translation of end-user requirements into explanations supported by AI systems. We are taking a two-pronged approach to pursue future work in operationalizing the use of our explanation ontology for clinical decision support systems. To this end, we are working towards building an explanation as a service that would leverage our explanation ontology and connect with AI methods to support the generation of components necessary to populate explanation types in different use cases. Further, we are implementing the clinical prototype as a functional UI with affordances for explanations. We expect the guidance along with open-sourced resources on our website will be useful to system designers looking to utilize our ontology to model and plan for explanations to include in their systems. 

\section*{Acknowledgments}

This work is done as part of the HEALS
project, and is partially supported by IBM Research AI through the AI Horizons Network. 

\bibliographystyle{splncs04}
\bibliography{references}   
\end{document}